\documentclass{article}

\PassOptionsToPackage{numbers}{natbib}
\usepackage[preprint, nonatbib]{neurips_2020}
\usepackage[utf8]{inputenc} % allow utf-8 input
\usepackage[T1]{fontenc}    % use 8-bit T1 fonts
\usepackage{hyperref}       % hyperlinks
\usepackage{url}            % simple URL typesetting
\usepackage{booktabs}       % professional-quality tables
\usepackage{amsfonts}       % blackboard math symbols
\usepackage{nicefrac}       % compact symbols for 1/2, etc.
\usepackage{microtype}      % microtypography
\usepackage{graphicx}
\usepackage{amsmath}

\title{Word2vec Skip-gram Dimensionality Selection via Sequential Normalized Maximum Likelihood}

\author{
  Pham Thuc Hung \\
  The Department of Creative Informatics \\
  Graduate School of Information Science \\ and Technology \\
  The University of Tokyo \\
  Hongo, Bunkyo-ku, Japan \\
  \texttt{hung.pham.thuc@ci.i.u-tokyo.ac.jp} \\
  \And Kenji Yamanishi \\
  The Department of Mathematical Informatics \\
  Graduate School of Information Science \\ and Technology \\
  The University of Tokyo \\
  Hongo, Bunkyo-ku, Japan \\
  \texttt{yamanishi@mist.i.u-tokyo.ac.jp} \\
}

\begin{document}

\maketitle

\begin{abstract}
  In this paper, we propose a novel information criteria-based approach to select the dimensionality of the word2vec Skip-gram (SG). From the perspective of the probability theory, SG is considered as an implicit probability distribution estimation under the assumption that there exists a true contextual distribution among words. Therefore, we apply information criteria  with the aim of selecting the best dimensionality so that the corresponding model can be as close as possible to the true distribution. We examine the following information criteria for the dimensionality selection problem: the Akaike’s Information Criterion, Bayesian Information Criterion, and Sequential Normalized Maximum Likelihood (SNML) criterion. SNML is the total codelength required for the sequential encoding of a data sequence on the basis of the minimum description length. The proposed approach is applied to both the original SG model and the SG Negative Sampling model to clarify the idea of using information criteria. Additionally, as the original SNML suffers from computational disadvantages, we introduce novel heuristics for its efficient computation. Moreover, we empirically demonstrate that SNML outperforms both BIC and AIC. In comparison with other evaluation methods for word embedding, the dimensionality selected by SNML is significantly closer to the optimal dimensionality obtained by word analogy or word similarity tasks.
\end{abstract}

\section{Introduction}
    In recent years, word2vec has been widely applied to many aspects of Natural Language Processing (NLP) and information retrieval such as machine translation \cite{luong-etal-2015-effective, NIPS2017_7181}, text classification \cite{conf/IEEEicci/LillebergZZ15}, text summarization \cite{nallapati-etal-2016-abstractive}, and named entity recognition \cite{Sien}. Furthermore, word2vec is used  in various fields such as materials science \cite{31849}, healthcare \cite{Gligorijevic}, and recommendation engines \cite{barkan2016item2vec, 10.1145/3219819.3219885, 10.1145/3219819.3219869}.
    
    The selection of the dimensionality for word2vec is important with regard to two aspects: model accuracy and computing resources. It is crucial to have a model of dimensionality high enough to learn the regularity of the data, but too high a dimensionality tends to cause overfitting. In addition, a large model is accompanied by a massive number of parameters for storage in the machine during training \cite{shu2018compressing}, leading to wasted memory resources. Thus, it is crucial to devise a method that can decide upon a dimensionality that satisfies the ability to capture necessary information from training data as well as makes efficient use of the computational resources.
    
    However, few studies have focused on the dimensionality selection problem. Most research evaluating the effectiveness of word embedding focuses on word analogy and word similarity tasks \cite{10.5555/3016387.3016557}. These evaluation methods require handcrafted datasets for implementation, but such datasets are currently not available to evaluate model training on non-English verbal and non-verbal data. To the best of our knowledge, only Yin and Shen \cite{yin2018dimensionality} accomplished the dimensionality selection of a word embedding model without the use of evaluation datasets. However, two aspects about this method need further consideration: the assumption that the noise signal obeys the zero mean-Gaussian distribution has not been verified in real data, and the selected dimensionality is quite different from those obtained by the other evaluation methods based on handcrafted datasets.
    
    From the perspective of information theory, we introduce an information criteria-based approach. We specifically propose the Sequential Normalized Maximum Likelihood (SNML) criterion in combination with some heuristics for the dimensionality selection problem. Application of our proposed criterion ensures that the selected dimensionality is able to capture regularity from the data as well as meet the preferences of models with relatively low but sufficient dimensionality. This information criteria-based approach does not require handcrafted evaluation datasets. To the best of our knowledge, this study presents the first application of information criteria in the field of word presentation as well as the first heuristic comparison of the SNML codelength. The results of the experiments show that the dimensionality selected by our method provides appropriate performance compared to optimal dimensionality for word analogy and word similarity tasks for English text data, and the estimated contextual distribution using this dimensionality is the closest thus far to the true distribution generated in synthetic data.

\section{Related work}
\subsection{Word embedding}
    Representations of words in a vector space have been studied exhaustively in the NLP literature. Beginning with a one-hot vector (the very first representation of words), other word representation methods such as latent semantic analysis \cite{Deerwester1990IndexingBL} and latent Dirichlet allocation \cite{10.5555/944919.944937} have been proposed to improve NLP task performance over time. Various methods that represent words as dense vectors (referred to as “word embedding”), including GloVe \cite{Pennington14glove:global}, word2vec (SG and continuous bag of words) \cite{journals/corr/abs-1301-3781}, are considered as the state of the art  in this field. In this paper, we focus on SG, but the proposed approach can be applied to any other word embedding model.

    \paragraph{Formula of SG}
    As the SG model often uses the negative sampling technique, in this study, we work with both original SG (oSG) and Skip-gram with Negative Sampling (SGNS) to clarify the idea behind our approach. In order to apply information criteria on the SG, we summarize it and introduce our notations for both oSG and SGNS. Assume a corpus of words and their contexts: $\mathcal{D}=(\mathbf{w},\mathbf{c})=(w_1,c_1) (w_2,c_2 )  … (w_n,c_n )$; $w_i \in V_W, c_i \in V_C$, which are one-hot vectors, where $V_W$ and $V_C$ are the word and context vocabularies of sizes $S_W$ and $S_C$, respectively. The training process of oSG attempts to learn the contextual distribution for each word by maximizing the likelihood function seen below. 
    \begin{equation}
        P_{oSG}(\mathbf{c} | \mathbf{w};E,F) =
        \prod_{i=1}^{n} P_{oSG}(c_i | w_i;E,F) = 
        \prod_{i=1}^{n} \frac{\exp(w_i^T EF)c_i}
        {\sum_{c'\in\mathbb{C}} \exp(w_i^T EF)c'},
    \end{equation}
    where E and F are the parameter matrices of the shapes $(S_W \times d)$ and $(d \times S_C)$, respectively. $d$ is the dimensionality of the embedding vector space.
    
    Unlike oSG, SGNS learns the probability that a particular context occurred around a word or not: $P(x_{i0} = 1|w_i,c_i;E,F)$. Furthermore, SGNS introduces $S_z$ negative samples $z_i=\{z_{i1}, z_{i2}, ..., z_{iS_z}\} \in V_C^{(S_z)} $ for each particular word $w_i$: $P(x_{ij}=0|w_i,z_{ij};E,F); j=\{1, 2, ..., S_z\}$. The training process of SGNS attempts to maximize the following likelihood function: 
    \begin{equation}
        P_{SGNS}(\mathbf{x}|\mathbf{w},\mathbf{c},\mathbf{z};E,F)=
        \prod_{i=1}^{n}{P_{SGNS}(x_i|w_i,c_i,z_i;E,F)},
    \end{equation}
    \begin{equation}
    \begin{split}
        P_{SGNS}(x_i|c_i,z_i,w_i;E,F) & =
        P_{SGNS}(x_{i0}=1|w_i,c_i;E,F)\prod_{j=1}^{S_Z}{P_{SGNS}(x_{ij}=0|w_i,z_{ij};E,F)} \\ & =\sigma({w_i}^T EFc_i)\prod_{j=1}^{S_Z}{\sigma(-{w_i}^T EFz_{ij})},
    \end{split}
    \end{equation}
    In the remainder of this paper, we denote $P(\mathcal{D};\theta)$ for both $P_{oSG}(\mathbf{c} | \mathbf{w};E,F)$ and $P_{SGNS}(\mathbf{x}|\mathbf{w},\mathbf{c},\mathbf{z};E,F)$.
    
\subsection{Dimensionality of SG}
    Unlike our approach, Yin and Shen \cite{yin2018dimensionality} considered word embedding to be an implicit matrix factorization problem \cite{10.5555/2969033.2969070}, and approached the issue by deciding the rank of the component matrix. Their work was conducted by introducing Pairwise Inner Product (PIP) loss, a measure that evaluates the goodness of the rank of matrix factorization. The best rank is chosen to minimize a given upper bound of the PIP loss.
    
    However, the selected number of dimensions does not agree with the optimal dimensionality performance based on the other evaluation tasks. For example, the best dimensionality of SG chosen by PIP loss is \textbf{129}, and the best 5\% dimensionalities range from 67 to 218, while the best dimensionalities in the WordSim353 (WS), MTurk771 (MTurk), and Google word analogy (WA) datasets are \textbf{56}, \textbf{102}, and \textbf{220}, respectively \cite{yin2018dimensionality}. Moreover, the matrix factorization operation conducted during the PIP loss calculation suffers from computational disadvantages and exceeds the calculation limit for huge amounts of data (e.g. Wikipedia dataset in our experiments).

\subsection{Information criteria}
    Word2vec is classified as a self-supervised machine learning model. Therefore, the number of dimensions can be selected by comparing the value of the loss function on the validation dataset. An alternative approach to dimensionality selection involves using information criteria such as the Akaike Information Criterion (AIC) \cite{akaike1973information}, Bayesian Information Criterion (BIC) \cite{Schwarz_1978}, and Minimum Description Length (MDL)  \cite{10.1016/0005-1098(78)90005-5}. Compared to the cross-validation method, these information criteria do not require a hold-out validation dataset, which prevents wastage of our precious data.
    
    Since AIC, BIC, and MDL have different backgrounds with regard to the estimation of expected log likelihood and approximation of the log marginal likelihood, we need to carefully choose the criteria to be used in specific cases. In fact, AIC and BIC rely heavily on the asymptotic theory, which states that as the data size grows to infinity, the estimated parameters converge in probability to the true values of the parameters. However, the asymptotic theory does not apply to word2vec; i.e. as the number of data increases to infinity, we can obtain different optimal parameters set (E, F). Therefore, AIC and BIC are not guaranteed to work theoretically. Nonetheless, several empirical studies have applied them successfully.
    
    Unlike AIC and BIC, MDL with Normalized Maximum Likelihood (NML) codelength is an accurate model selection criterion for real-world data analysis based on limited samples. NML is also known as the best codelength in the context of the minimax optimality property \cite{Sht87}. 
    
    However, choosing the best method for dimensionality selection is still an experimental task in word2vec. In the next section, we describe in detail the application of MDL to the dimension selection problem and the reason for choosing this method. We then provide empirical comparisons between the methods listed in this section.
    
    \paragraph{Formulae for AIC and BIC}
    In order to apply these information criteria to the dimensionality selection problem, we introduce our notations for the AIC and BIC first.
    \begin{equation}
    AIC=2 (S_W\times d+d\times S_C)-2\ln{\left(P(\mathcal{D};\hat{\theta}(\mathcal{D}))\right)},
    \end{equation}
    \begin{equation}
    BIC=\ln{\left(n\right)} (S_W\times d+d\times S_C) - 2\ln{\left(P(\mathcal{D};\hat{\theta}(\mathcal{D}))\right)},
    \end{equation}
    where, $\hat{\theta}\left(\mathcal{D}\right)=(\hat{E}\left(\mathcal{D}\right),\hat{F}\left(\mathcal{D}\right))$ is the maximum likelihood estimation of the parameters on data $\mathcal{D}$.

\section{Dimensionality selection via the MDL principle}
\subsection{Applying the MDL principle, NML and SNML codelengths}
    Word2vec was derived based on the distributional hypothesis of Harris \cite{harris1954distributional}, which states that words in similar contexts have similar meanings. Therefore, assuming the existence of the true context distribution for given words $P^\ast(\cdot|w)$, it is reasonable to choose the dimensionality that has the ability to learn the context distribution most similar to the true distribution. The MDL principle \cite{10.1016/0005-1098(78)90005-5} is a powerful solution for model selection, and is considered for the dimensionality selection as per our interest.
    
    The MDL principle states that the best hypothesis (a model and its parameters) for a given set of data is the one that leads to the best compression of the data, namely the minimum codelength \cite{10.1016/0005-1098(78)90005-5}. Specifically, we consider each dimensionality corresponding to a probability model class $\mathcal{M}_d$.
    \begin{equation}
        \mathcal{M}_d= \{ P\left(\mathcal{D};\theta\right):\ \theta = (E\in R^{\left(S_W\times d\right)},\ F\in R^{\left(d \times S_C\right)} ) \},
    \end{equation}
    We take the expression $\mathcal{L}\left( \mathcal{D} ;\mathcal{M}_d\right)$ as the codelength of data $\mathcal{D}$ that can be obtained for the given model class $\mathcal{M}_d$. Therein, the NML  codelength is the best-known codelength in the MDL literature to achieve the minimax regret \cite{Sht87}. The formula for the NML codelength is given below. 
    \begin{equation}
        \mathcal{L}_{NML} \left( \mathcal{D} ; \mathcal{M}_d \right) = - \log{P ( \mathcal{D};\hat{\theta}(\mathcal{D}))}+\log{C(\mathcal{M}_d)},
    \end{equation}
    where $\log{C(\mathcal{M}_d)}=\log{\sum_{\mathcal{D}\in \mathbb{D}^{(n)}}{P( \mathcal{D};\hat{\theta}(\mathcal{D}))}}$ is known as Parametric Complexity (PC); $\mathbb{D}^{(n)}$ denotes for all possible data series with the length of $n$.
    
    However, the PC term involves extensive computations and is not realistic to implement. Instead, we apply the SNML codelength \cite{RISSANEN2010839} in this study to reduce the computation cost using the formula seen below: 
    \begin{equation}
        \mathcal{L}_{SNML}( \mathcal{D} ;\mathcal{M}_d)= \sum_{i=1}^{n}{\mathcal{L}_{SNML} ( \mathcal{D}_i| \mathcal{D}^{i-1} ;\mathcal{M}_d)},
    \end{equation}
    where $\mathcal{D}^i$ denote for data series $\mathcal{D}_1,\mathcal{D}_2,...,\mathcal{D}_i$ and $\mathcal{D}=\mathcal{D}_1,\mathcal{D}_2,...,\mathcal{D}_n$. The SNML codelength is calculated as the total codelength where the codelength for each data is sequentially calculated such as the NML codelength every time it is input. It is known that the SNML codelength is a good approximation to the NML codelength \cite{rissanen_2012}. Since the SNML codelength is sequentially calculated, its computational cost at each step is much lower than that of the NML codelength.
    
    In addition, the SNML codelength function $\mathcal{L}_{SNML}( \mathcal{D}_i| \mathcal{D}^{i-1} ;\mathcal{M}_d)$ can be applied to oGS and SGNS in the forms seen below. 
    \begin{equation}
    \begin{split}
        \mathcal{L}_{SNML}(\mathcal{D}_i |\mathcal{D}^{i-1};\mathcal{M}^{oSG}_d) & = -\log{P_{oSG}(c_i|w^i,c^{i-1};\hat{\theta}(w^i,c^i))} \\ 
        & + \log{\sum_{c\in V_C}{P_{oSG}(c|w^i,c^{i-1};\hat{\theta}(w^i,c^{i-1},c))}},
    \end{split}
    \end{equation}
    \begin{equation}
    \begin{split}
        \mathcal{L}_{SNML}(\mathcal{D}_i |\mathcal{D}^{i-1};\mathcal{M}^{SGNS}_d) = -\log{P_{SGNS}(x_i|w^i,c^i,z^i,x^{i-1};\hat{\theta}(w^i,c^i,z^i,x^i))} \\ + \log{\sum_{x\in \mathcal{O}^{(S_z)}}{P_{SGNS}(x|w^i,c^i,z^i,x^{i-1};\hat{\theta}(w^i,c^i,z^i,x^{i-1},x))}},
    \end{split}
    \end{equation}
    where $\mathcal{O}^{(S_z)}$ is set of all possible one-hot vector of $S_z$ dimensions. 

\subsection{Some heuristics associated with SNML codelength calculation}
    The computation of the SNML codelength still costs $nS_C$ times to train each data record $\mathcal{D}_i$, which is also not realistic. We introduce two techniques for saving the computational costs for SNML: heuristic comparison and importance sampling on the SNML codelength.

\paragraph{Heuristic comparison}
    A simple observation reveals that if the codelength of data obtained with model class $\mathcal{M}_d$ is the shortest, then only some part of the data can also be achieved with the shortest codelength compressed with the same model class. Therefore, instead of computing the codelength for all $n$ records of data, we can use the codelength of a small set of data. In fact, the results of our experiments show that focusing on the last several thousand records of data are sufficient to compare model classes.  
    
\begin{figure}
  \centering
  \includegraphics[scale=0.45]{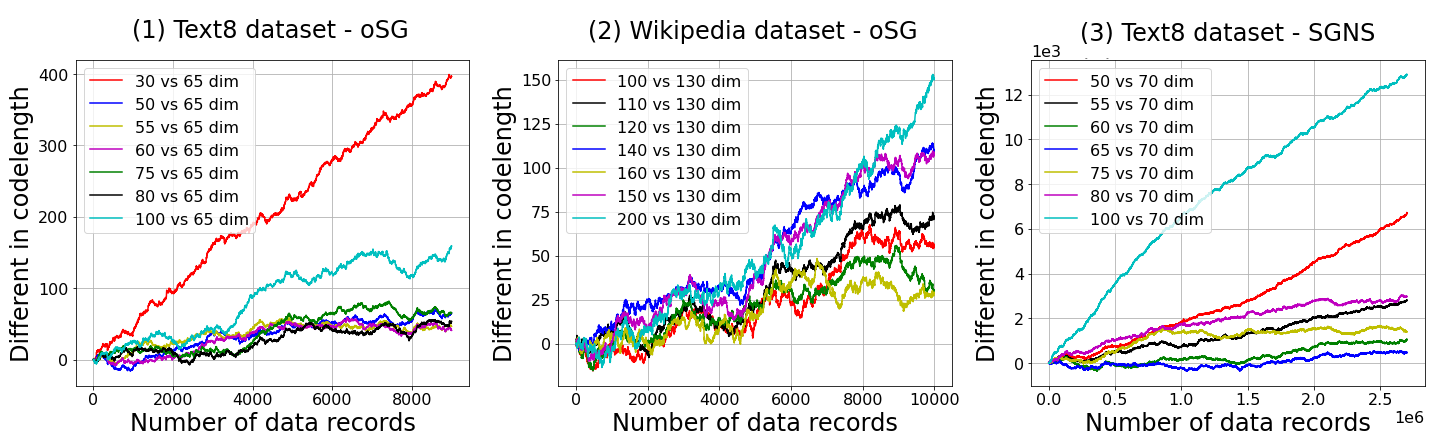}
  \caption{Cumulative SNML codelengths of different dimensionalities compared to the dimensionality result with the shortest codelength}
\end{figure}
    Figure 1 demonstrates the differences in SNML codelengths of different dimensionalities compared with the dimensionality that achieves the shortest codelength on the data. The vertical axis show the difference of data codelength obtained by two different dimensionality shown in the legends (e.g. $d1\ vs\ d2\ dim$); specifically, it is calculated by $\mathcal{L}(\mathcal{D'}; d1)-\mathcal{L}( \mathcal{D'}; d2)$ where $\mathcal{L}$ is the codelength function; $\mathcal{D'}$ is data; $d1$ and $d2$ are dimensionality; while the value of horizontal axis shows the number of records of $\mathcal{D'}$.
    
    To facilitate comparisons among dimensionalities that are markedly different from one another (such as 30 dimensions versus 65 dimensions in Figure 1-1, or 200 dimensions versus 130 dimensions in Figure 1-2), it is sufficient to use only 6,000 data records to provide information about the best dimensionality to be chosen. Therefore, adding data thereafter simply increases the SNML codelength but does not change our answer substantially. However, for similar dimensionalities, such as 60, 65, and 70 dimensions in Figure 1-3, the first one million data records cannot help us identify the optimal dimensionality. This phenomenon leads to confusion when the codelengths between two model classes are not too different. To ensure that the correct model is chosen, we need to increase the number of records to estimate the SNML codelength so as to allow a better comparison of these two dimensionalities. However, a small dimension error in the dimensionality selection of word2vec does not affect the final performance considerably. Therefore, the trade-off between the computing time and model selection accuracy is determined by the number of records beyond those required to estimate the SNML codelength with sufficient finality.
    
\paragraph{Importance sampling}
    Since the size of the context set $S_C$ is large (approximately 30,000–100,000 or above, according to the training dataset), the computation of PC for SNML in oSG is still very expensive. We apply the importance sampling method to approximately estimate the SNML description length for each data record. In detail, if a distribution $Q$ on the context set satisfies $Q\left(c\right)\neq0\ \forall\ c\in V_C$, the following formula can be applied.
    
    Let $f\left(c\right)=\ P\left(c\middle| w^i,\ c^{i-1};\hat{\theta}(w^i,c^{i-1},c)\right)$, then
    \begin{equation}
        \sum_{j=1}^{S_C}f\left(c_j\right)=\sum_{j=1}^{S_C}{Q\left(c_j\right)\frac{f\left(c_j\right)}{Q\left(c_j\right)}}=\mathbb{E}_Q\left(\frac{f(c)}{Q(c)}\right)\approx\frac{1}{m}\sum_{c\in\mathcal{S}}\frac{f(c)}{Q(c)},
    \end{equation}
    where $\mathcal{S}=\left\{c_1,\ c_2,..,\ c_m\right\}\ \sim \ Q(c)$: set of samples draw from distribution Q.
    
    This estimation asymptotes to the true value as $m$ (the number of samples) increases, and distribution $Q$ is similar to function $f(c)$. In our experiment, the uniform distribution is the best choice for distribution $Q$, and the sampling size is chosen to be 1/10 the size of the context set to balance the computation time and sampling error.

\section{Experimental settings}
\subsection{Data}
    We compare the above-mentioned model selection criteria using SG trained on three datasets: synthetic data, text8, and Wikipedia.

\paragraph{Synthetic data}
    Synthetic data are generated based on several random questions from the WA dataset. Assuming a numeric context set, we generate categorical distributions on this set for all words for which the parameter vectors of the corresponding distributions satisfy the constraints in the questions. For example, corresponding to question: $Tokyo, Japan, Paris, France$, the process involves generation of four random contextual distributions, $\widetilde{P}(\cdot| T o k y o)$, $\widetilde{P}(\cdot| J a p a n)$, $\widetilde{P}(\cdot| P a r i s)$, $\widetilde{P}(\cdot| F r a n c e)$, such that:
    $cosine(\widetilde{P}(\cdot| T o k y o), \widetilde{P}(\cdot| J a p a n)) = cosine(\widetilde{P}(\cdot| P a r i s), \widetilde{P}(\cdot| F r a n c e))$. The implementation for generation of such categorical distributions is also available on GitHub\footnote{Currently unavailable in submission version}.
    
    We then sample words using a uniform distribution and contexts using $\widetilde{P}$ adding normal distribution noises. Using these pairs of word and context, oSG and SGNS can be trained to achieve a 100\% score on the questions used to create data with the appropriate dimensionality. Furthermore, good dimensionality should result in contextual distributions similar to $\widetilde{P}$. To evaluate this similarity, we use a dissimilar function for the oSG model and a similar function for the SGNS model as follows:
    \begin{equation}
        dissimilar(\mathcal{M}^{(oSG)}_d, \widetilde{P}) = 
        \frac{1}{S_W}\sum_{w\in V_W}{D_{KL}(P_{oSG}(\cdot|w;\hat{\theta})||\widetilde{P}(\cdot|w))},
    \end{equation}
    \begin{equation}
        similar(\mathcal{M}^{(SGNS)}_d, \widetilde{P}) = 
        \frac{1}{S_W}\sum_{w\in V_W}{\rho (f_{P_{SGNS}}(\cdot|w;\hat{\theta}), f_{\widetilde{P}}(\cdot|w))},
    \end{equation}
    where, $D_{KL}$ denotes for Kullback–Leibler divergence, $\rho$ denotes Spearman's rank correlation coefficient, $f_{P_{SGNS}}(\cdot|w;\hat{\theta})$ and $f_{\widetilde{P}}(\cdot|w)$ are vectors that take $P_{SGNS}(x=1|w,c;\hat{\theta})$ and $\widetilde{P}(c|w)$  ($c \in V_C$) as elements, respectively. The choice of $D_{KL}$ for oSG comes from the fact that oSG outputs a categorical distribution; while SGNS results a list of probability values which are expected to have a strong positive correlation with values of $\widetilde{P}$. We use $dissimilar(\mathcal{M}^{(oSG)}_d, \widetilde{P})$ and $similar(\mathcal{M}^{(SGNS)}_d, \widetilde{P})$ as the oracle criterion to evaluate the optimal dimensionality for synthetic data.

\paragraph{Text datasets}
    The text8 and Wikipedia datasets are preprocessed using a window size of 5, removing words that occur less than 73 times, and applying subsampling with a threshold of ${10}^{-5}$. In addition, we only use the first 20,000 articles of the English Wikipedia dump  for the training process.

\subsection{Training process}
\paragraph{Optimization settings}
    In order to speed up the training process, we implement momentum optimizer and mini-batch with a batch size of 1,000 for oSG training and stochastic gradient descent for SGNS, as in \cite{NIPS2013_5021}. Learning rate $\alpha$ for oSG is set to 1.0 and momentum is 0.9. For SGNS, $\alpha$ is choosen to be 0.1, and the number of negative samples is 15. The number of epochs is chosen so that the negative log likelihood value is not significantly reduced. For instance, in the case of oSG, 200 and 90 epochs respectively were selected for text8 and Wikipedia, while for SGNS, 15 epochs were selected for text8.  Practically, these optimization settings achieve the best performance in our experiment. For example, the best word analogy (WA) scores for text8 are 32.6\% (SGNS) and 38.6\% (oGS), the corresponding value for Wikipedia is 50.5\% (oNS).
    
    Because of the limitations posed by the computational resources, we experiment on a finite number of dimensionalities, which we think is sufficient to clarify the idea behind this research. The evaluated dimensionalities are then shown in the figures corresponding to each dataset.

\paragraph{Estimation of SNML codelength}
    Estimation of SNML codelength requires us to repeat the parameter estimation $\hat{\theta}(w^i,c^i)$ $s\times m$ times, where s is the number of records beyond those required to estimate the SNML codelength, and m is the sampling size. However, repeatedly estimating parameters from scratch is very time consuming. We can alternatively estimate $\hat{\theta}(w^i,c^i)$ from $\hat{\theta}(w^{i-1},c^{i-1})$ by taking the gradient descent of $(w_i,w_i)$.

\section{Experimental results}
\subsection{Synthetic data}
    We compare five criteria: AIC, BIC, SNML, accuracy on the WA task, and loss value on the validation dataset (CV) with the oracle criterion. The experimental results are shown in Figure 2. Due to the differences between the criteria values, we scale all the values to range from 0 to 1 for visual purposes. Moreover, while the dissimilar oracle and other criteria take the dimensionality that minimizes the value, WA takes the maximum. Therefore, in the figure, we draw the line showing the negative value plus one for dissimilar oracle, AIC, BIC, SNML, and CV so that the higher value states better indicate the dimensionality to be chosen. This scale procedure is also applied to Figure 3 and Figure 4. The horizontal axis in these figures shows the number of dimensions.
    
\begin{figure}
  \centering
  \includegraphics[scale=0.65]{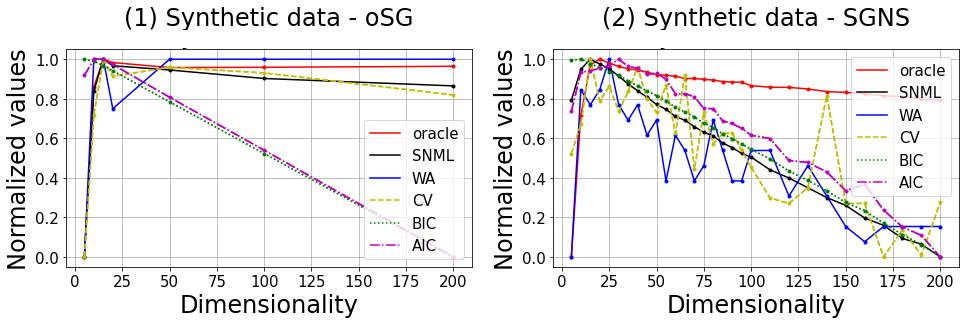}
  \caption{Normalized values of criteria compared with the oracle on synthetic data}
  
\end{figure}

    The results for oSG show that the BIC exhibits a monotonous decrease, while the AIC provides 10 as the best dimensionality. The results of the SNML and CV agree with the oracle, namely that 15 is the best dimensionality to choose. WA provides a result of 100\% over a long range. For SGNS, the oracle chooses 20 dimensions, SNML and CV choose 15, and WA achieves the highest score at 25 dimensions. On the other hand, the BIC chooses 10 dimensions while the AIC chooses 30.
    
    In both oSG and SGNS, SNML chooses the dimensionality equal or close to the oracle criterion, similar to CV. Thus, SNML outperforms both AIC and BIC. Note that the synthetic data are designed to achieve a 100\% WA score using contextual distribution; however, the scores achieved using embedded vectors change significantly according to the dimensionality; the optimal dimensionality for WA may be higher than that chosen by the oracle.

\subsection{Text data}
    We compare the SNML criterion with the NLP word analogy task (using WA) and word similarity tasks (using WS, MTurk, and MEN-3k test collection (MEN)). We experiment with at least 3 runs for each dataset, and the average results are shown in Figure 3. The comparison of SNML with the information criteria, CV, and PIP is depicted in Figure 4.
    
\begin{figure}
  \centering
  \includegraphics[scale=0.4]{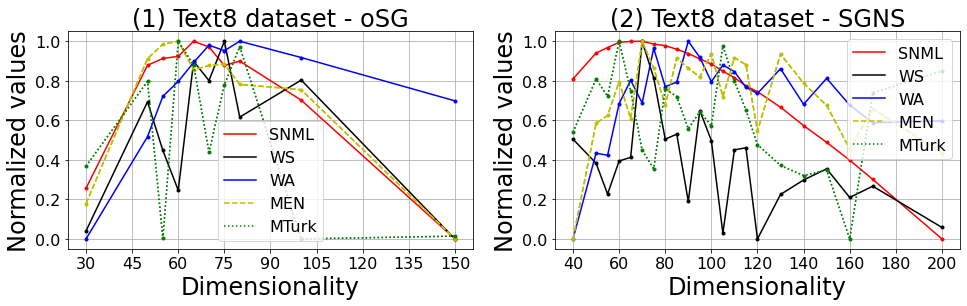}
  \caption{Normalized values and scores on NLP tasks compare with SNML}
\end{figure}

\begin{table}
  \caption{Optimal dimensionalities chosen by different criteria (-: unknown)}
  \label{sample-table}
  \centering
  \begin{tabular}{llllllllll}
    \toprule
                     & SNML& WS  & WA  & MEN & MTurk & CV  & AIC & BIC & PIP \\
    \midrule
    Text8 (oSG)      & 65  & 75  & 80  & 60  & 60    & 65  & -   & -   & 120 \\
    Wikipedia (oSG)  & 130 & 130 & 180 & 100 & 110   & 130 & 200 & -   & -   \\
    Text8 (SGNS)     & 70  & 70  & 95  & 70  & 60    & 60  & 70  & -   & 105 \\
    \bottomrule
  \end{tabular}
\end{table}
    
    As knowledge regarding the underlying true distribution of the data is lacking, it is difficult to determine the best measures. However, assuming the existence of the true contextual distribution, NLP tasks will roughly prioritize models closest to the true distribution. Therefore, the dimensionality selected by SNML is also close to the optimal dimensionalities for these tasks. The main results of the study are shown in the Table 1. In it, the optimal dimensionality chosen by the proposed method were compared with the optimal dimensionality in word analogy and word similarity tasks in NLP \cite{10.5555/3016387.3016557}. The closer the dimensionality of any method is to the optimal one in NLP task, the better the method is. Accordingly, for oSG, SNML and CV chose the same dimensionality which is closer to the optimal dimensionality in NLP tasks than AIC, BIC and PIP. For SGNS, SNML chose the dimensionality closer to the optimal dimensionality for 3 tasks (WS, WA, MEN) in 4 tasks (WA, WS, MEN and MTurk) implemented when compared to CV; and 4 in 4 tasks implemented when compared to BIC and PIP. We concluded that SNML is better than CV, AIC, BIC and PIP in almost implemented NLP tasks. Note that we are unable to implement PIP on Wikipedia because the computational complexity was beyond the capabilities of our servers. We are also unable to find the minimum values of BIC and AIC (for text8 train with oSG) for dimensions over a long range.
    
    Furthermore, the SNML criterion tends to favor smaller dimensions although it is sufficient to ensure good performance on other NLP tasks while heavily penalizing model classes that tend to overfitting. This characteristic of SNML helps us avoid choosing large models, and therefore, the available resources should be fully utilized.
    
\begin{figure}
  \centering
  \includegraphics[scale=0.4]{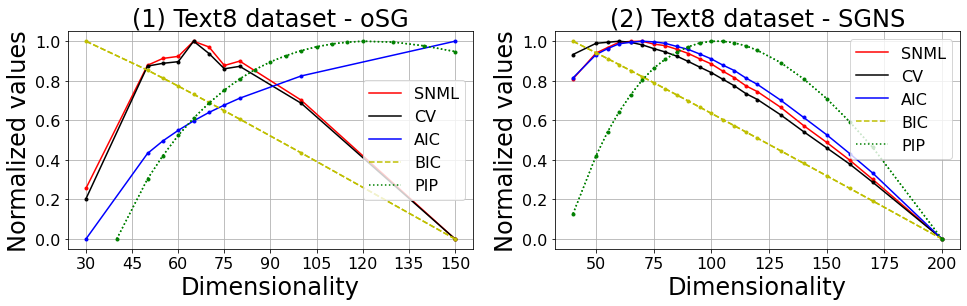}
  \caption{Normalized values of information criteria, CV, and PIP}
  
\end{figure}

\section{Conclusion}
    When considering word2vec SG as a probability distribution estimation problem, the optimal dimensionality can provide an estimation of contextual distribution as close as possible to the true distribution-generated data. We tested information criteria (AIC and BIC) and SNML with some heuristics to select such a dimensionality. The experimental results on synthetic data showed that the SNML could choose a dimensionality such that the corresponding probability model is able to learn the contextual distribution closest to the true distribution-generated data. The experiments on text datasets showed that SNML has the ability to choose a desirable dimensionality with regard to two aspects; low dimensionality although it is sufficient to ensure good performance in terms of significant closeness of optimal dimensionality in NLP tasks without a hold-out test dataset. Furthermore, SNML typically outperforms AIC, BIC, CV, and PIP in the selection of good dimensionality for NLP tasks in our experiments. Our method therefore holds promise for choosing the most appropriate dimensionality in word2vec when training with data is not limited to English or non-verbal data.
    
    To the best of our knowledge, this is the first study that applies information criteria to dimensionality selection for word embedding. In fact, the limitations associated with computation or asymptotic estimation of NML or SNML codelength make it difficult to apply such criteria in these areas. By introducing some heuristics in the SNML codelength calculation, we have discovered a new and useful approach, namely MDL-based knowledge embedding. Our proposed approach can be applied to other embedding methods once a likelihood function corresponding to any embedding method is defined. For example, in the case of BERT, the likelihood function can be defined using a joint probability distribution of masked token  and next sentence. This likelihood function is then substituted for distribution $P$ in Eqs. (6), (7), (8)  to obtain NML and SNML codelengths. More detailed evaluation will be left for future study.

\section*{Broader Impact}
    
    This paper makes two main contributions to the literature: the presented method effectively estimates the optimal dimensionality for word2vec models as well as successfully compares SNML codelength between model classes. Estimation of the optimal dimensionality for word2vec models can help researchers and engineers improve the performance of artificial intelligence (AI) systems using word2vec. When the dimensionality selected by the method fails, the performance of the AI system degrades. Our proposed approach can help avoid such failures. Broadly speaking, users and stakeholders of such systems may benefit from advances in AI performance. For example, improvements in the recommendation engines using word2vec will lead to a better user experience, and the system owner may also gain a higher conversion rate or revenue. 
    
    Our method of comparing the SNML codelength among model classes introduces a new research direction about hyper-parameters for other embedding methods or neural networks models. In the case of deep learning (DL) models, due to the preference to lower dimensionality of AIC, BIC or SNML, proposed method does not benefit from the over-parameterization property of these models. However, other improvements such as considering an effective parameterization method for AIC and BIC can also be straightforwardly applied to our framework in order to help selected dimensionality benefit from the over-parameterization property of DL models. It will be a future challenge in this field to investigate what improvement yields the best strategy for hyper-parameters selection.

\medskip

\small
\bibliography{neurips_2020}

\begin{thebibliography}{26}
\providecommand{\natexlab}[1]{#1}
\providecommand{\url}[1]{\texttt{#1}}
\expandafter\ifx\csname urlstyle\endcsname\relax
  \providecommand{\doi}[1]{doi: #1}\else
  \providecommand{\doi}{doi: \begingroup \urlstyle{rm}\Url}\fi

\bibitem[Luong et~al.(2015)]{luong-etal-2015-effective}
Thang Luong, Hieu Pham, and Christopher~D. Manning.
\newblock Effective approaches to attention-based neural machine translation.
\newblock In \emph{Proceedings of the 2015 Conference on Empirical Methods in
  Natural Language Processing}, pages 1412--1421, Lisbon, Portugal, September
  2015. Association for Computational Linguistics.
\newblock \doi{10.18653/v1/D15-1166}.
\newblock URL \url{https://www.aclweb.org/anthology/D15-1166}.

\bibitem[Vaswani et~al.(2017)]{NIPS2017_7181}
Ashish Vaswani, Noam Shazeer, Niki Parmar, Jakob Uszkoreit, Llion Jones,
  Aidan~N Gomez, \L~ukasz Kaiser, and Illia Polosukhin.
\newblock Attention is all you need.
\newblock In I.~Guyon, U.~V. Luxburg, S.~Bengio, H.~Wallach, R.~Fergus,
  S.~Vishwanathan, and R.~Garnett, editors, \emph{Advances in Neural
  Information Processing Systems 30}, pages 5998--6008. Curran Associates,
  Inc., 2017.
\newblock URL
  \url{http://papers.nips.cc/paper/7181-attention-is-all-you-need.pdf}.

\bibitem[Lilleberg et~al.(2015)]{conf/IEEEicci/LillebergZZ15}
Joseph Lilleberg, Yun Zhu, and Yanqing Zhang.
\newblock Support vector machines and word2vec for text classification with
  semantic features.
\newblock In Ning Ge, Jianhua Lu, Yingxu Wang, Newton Howard, Philip Chen,
  Xiaoming Tao, Bo~Zhang, and Lotfi~A. Zadeh, editors, \emph{ICCI*CC}, pages
  136--140. IEEE Computer Society, 2015.
\newblock ISBN 978-1-4673-7290-9.
\newblock URL
  \url{http://dblp.uni-trier.de/db/conf/IEEEicci/IEEEicci2015.html#LillebergZZ15}.

\bibitem[Nallapati et~al.(2016)]{nallapati-etal-2016-abstractive}
Ramesh Nallapati, Bowen Zhou, Cicero dos Santos, {\c{C}}a{\u{g}}lar
  GuÌ‡l{\c{c}}ehre, and Bing Xiang.
\newblock Abstractive text summarization using sequence-to-sequence {RNN}s and
  beyond.
\newblock In \emph{Proceedings of The 20th {SIGNLL} Conference on Computational
  Natural Language Learning}, pages 280--290, Berlin, Germany, August 2016.
  Association for Computational Linguistics.
\newblock \doi{10.18653/v1/K16-1028}.
\newblock URL \url{https://www.aclweb.org/anthology/K16-1028}.

\bibitem[Sienčnik(2015)]{Sien}
Scharolta~Katharina Sienčnik.
\newblock Adapting word2vec to named entity recognition.
\newblock In \emph{Proceedings of the 20th Nordic Conference of Computational
  Linguistics, NODALIDA 2015, May 11-13, 2015, Vilnius, Lithuania}, number 109,
  pages 239--243. Linköping University Electronic Press, Linköpings
  universitet, 2015.

\bibitem[Tshitoyan et~al.(2019)]{31849}
Vahe Tshitoyan, John Dagdelen, Leigh Weston, Alexander Dunn, Ziqin Rong, Olga
  Kononova, Kristin~A. Persson, Gerbrand Ceder, and Anubhav Jain.
\newblock Unsupervised word embeddings capture latent knowledge from materials
  science literature.
\newblock \emph{Nature}, 571:\penalty0 95 -- 98, 07/2019 2019.
\newblock ISSN 0028-0836.
\newblock \doi{10.1038/s41586-019-1335-8}.

\bibitem[Gligorijevic et~al.(2016)]{Gligorijevic}
Djordje Gligorijevic, Jelena Stojanovic, Nemanja Djuric, Vladan Radosavljevic,
  Mihajlo Grbovic, Rob~J. Kulathinal, and Zoran Obradovica.
\newblock Large-scale discovery of disease-disease and disease-gene
  associations.
\newblock \emph{Scientific report}, 6 32404, 08/2016 2016.
\newblock \doi{10.1038/srep32404}.

\bibitem[Barkan and Koenigstein(2016)]{barkan2016item2vec}
Oren Barkan and Noam Koenigstein.
\newblock Item2vec: Neural item embedding for collaborative filtering, 2016.
\newblock URL \url{http://arxiv.org/abs/1603.04259}.
\newblock cite arxiv:1603.04259.

\bibitem[Grbovic and Cheng(2018)]{10.1145/3219819.3219885}
Mihajlo Grbovic and Haibin Cheng.
\newblock Real-time personalization using embeddings for search ranking at
  airbnb.
\newblock In \emph{Proceedings of the 24th ACM SIGKDD International Conference
  on Knowledge Discovery \& Data Mining}, KDD ’18, page 311–320, New York,
  NY, USA, 2018. Association for Computing Machinery.
\newblock ISBN 9781450355520.
\newblock \doi{10.1145/3219819.3219885}.
\newblock URL \url{https://doi.org/10.1145/3219819.3219885}.

\bibitem[Wang et~al.(2018)]{10.1145/3219819.3219869}
Jizhe Wang, Pipei Huang, Huan Zhao, Zhibo Zhang, Binqiang Zhao, and Dik~Lun
  Lee.
\newblock Billion-scale commodity embedding for e-commerce recommendation in
  alibaba.
\newblock In \emph{Proceedings of the 24th ACM SIGKDD International Conference
  on Knowledge Discovery \& Data Mining}, KDD ’18, page 839–848, New York,
  NY, USA, 2018. Association for Computing Machinery.
\newblock ISBN 9781450355520.
\newblock \doi{10.1145/3219819.3219869}.
\newblock URL \url{https://doi.org/10.1145/3219819.3219869}.

\bibitem[Shu and Nakayama(2018)]{shu2018compressing}
Raphael Shu and Hideki Nakayama.
\newblock Compressing word embeddings via deep compositional code learning.
\newblock In \emph{International Conference on Learning Representations}, 2018.
\newblock URL \url{https://openreview.net/forum?id=BJRZzFlRb}.

\bibitem[Zhai et~al.(2016)]{10.5555/3016387.3016557}
Michael Zhai, Johnny Tan, and Jinho~D. Choi.
\newblock Intrinsic and extrinsic evaluations of word embeddings.
\newblock In \emph{Proceedings of the Thirtieth AAAI Conference on Artificial
  Intelligence}, AAAI’16, page 4282–4283. AAAI Press, 2016.

\bibitem[Yin and Shen(2018)]{yin2018dimensionality}
Zi~Yin and Yuanyuan Shen.
\newblock On the dimensionality of word embedding.
\newblock In S.~Bengio, H.~Wallach, H.~Larochelle, K.~Grauman, N.~Cesa-Bianchi,
  and R.~Garnett, editors, \emph{Advances in Neural Information Processing
  Systems 31}, pages 895--906. Curran Associates, Inc., 2018.
\newblock URL
  \url{http://papers.nips.cc/paper/7368-on-the-dimensionality-of-word-embedding.pdf}.

\bibitem[Deerwester et~al.(1990)]{Deerwester1990IndexingBL}
Scott~C. Deerwester, Susan~T. Dumais, Thomas~K. Landauer, George~W. Furnas, and
  Richard~A. Harshman.
\newblock Indexing by latent semantic analysis.
\newblock \emph{JASIS}, 41:\penalty0 391--407, 1990.

\bibitem[Blei et~al.(2003)]{10.5555/944919.944937}
David~M. Blei, Andrew~Y. Ng, and Michael~I. Jordan.
\newblock Latent dirichlet allocation.
\newblock \emph{J. Mach. Learn. Res.}, 3\penalty0 (null):\penalty0 993–1022,
  March 2003.
\newblock ISSN 1532-4435.

\bibitem[Pennington et~al.(2014)]{Pennington14glove:global}
Jeffrey Pennington, Richard Socher, and Christopher~D. Manning.
\newblock Glove: Global vectors for word representation.
\newblock In \emph{In EMNLP}, 2014.

\bibitem[Mikolov et~al.(2013{\natexlab{a}})]{journals/corr/abs-1301-3781}
Tomas Mikolov, Kai Chen, Greg Corrado, and Jeffrey Dean.
\newblock Efficient estimation of word representations in vector space.
\newblock \emph{CoRR}, abs/1301.3781, 2013{\natexlab{a}}.
\newblock URL
  \url{http://dblp.uni-trier.de/db/journals/corr/corr1301.html#abs-1301-3781}.

\bibitem[Mikolov et~al.(2013{\natexlab{b}})]{NIPS2013_5021}
Tomas Mikolov, Ilya Sutskever, Kai Chen, Greg~S Corrado, and Jeff Dean.
\newblock Distributed representations of words and phrases and their
  compositionality.
\newblock In C.~J.~C. Burges, L.~Bottou, M.~Welling, Z.~Ghahramani, and K.~Q.
  Weinberger, editors, \emph{Advances in Neural Information Processing Systems
  26}, pages 3111--3119. Curran Associates, Inc., 2013{\natexlab{b}}.
\newblock URL
  \url{http://papers.nips.cc/paper/5021-distributed-representations-of-words-and-phrases-and-their-compositionality.pdf}.

\bibitem[Levy and Goldberg(2014)]{10.5555/2969033.2969070}
Omer Levy and Yoav Goldberg.
\newblock Neural word embedding as implicit matrix factorization.
\newblock In \emph{Proceedings of the 27th International Conference on Neural
  Information Processing Systems - Volume 2}, NIPS’14, page 2177–2185,
  Cambridge, MA, USA, 2014. MIT Press.

\bibitem[Akaike(1973)]{akaike1973information}
Hirotogu Akaike.
\newblock \emph{Information Theory and an Extension of the Maximum Likelihood
  Principle}, pages 199--213.
\newblock Springer New York, New York, NY, 1973.

\bibitem[Schwarz(1978)]{Schwarz_1978}
Gideon Schwarz.
\newblock Estimating the dimension of a model.
\newblock \emph{The Annals of Statistics}, 6\penalty0 (2):\penalty0 461--464,
  March 1978.
\newblock \doi{10.1214/aos/1176344136}.
\newblock URL \url{https://doi.org/10.1214\%2Faos\%2F1176344136}.

\bibitem[Rissanen(1978)]{10.1016/0005-1098(78)90005-5}
Jorma Rissanen.
\newblock Paper: Modeling by shortest data description.
\newblock \emph{Automatica}, 14\penalty0 (5):\penalty0 465–471, September
  1978.
\newblock ISSN 0005-1098.
\newblock \doi{10.1016/0005-1098(78)90005-5}.
\newblock URL \url{https://doi.org/10.1016/0005-1098(78)90005-5}.

\bibitem[Shtarkov(1987)]{Sht87}
Yu~M. Shtarkov.
\newblock Universal sequential coding of single messages.
\newblock \emph{Problems Information Transmission}, 23\penalty0 (3):\penalty0
  3--17, 1987.

\bibitem[Harris(1954)]{harris1954distributional}
Zellig~S Harris.
\newblock Distributional structure.
\newblock \emph{Word}, 10\penalty0 (2-3):\penalty0 146--162, 1954.

\bibitem[Rissanen et~al.(2010)]{RISSANEN2010839}
Jorma Rissanen, Teemu Roos, and Petri Myllymäki.
\newblock Model selection by sequentially normalized least squares.
\newblock \emph{Journal of Multivariate Analysis}, 101\penalty0 (4):\penalty0
  839 -- 849, 2010.
\newblock ISSN 0047-259X.
\newblock \doi{https://doi.org/10.1016/j.jmva.2009.12.009}.
\newblock URL
  \url{http://www.sciencedirect.com/science/article/pii/S0047259X09002401}.

\bibitem[Rissanen(2012)]{rissanen_2012}
Jorma Rissanen.
\newblock \emph{Optimal Estimation of Parameters}.
\newblock Cambridge University Press, 2012.
\newblock \doi{10.1017/CBO9780511791635}.

\end{thebibliography}

\end{document}